\documentclass[letterpaper, 10 pt, conference]{ieeeconf}  % Comment this line out if you need a4paper

\IEEEoverridecommandlockouts                              % This command is only needed if 
\usepackage{graphics} % for pdf, bitmapped graphics files
\usepackage{tabularx}
\usepackage{amsmath, stmaryrd} % assumes amsmath package installed
\usepackage{cite}
\usepackage{diagbox}
\usepackage{amsmath,amssymb,amsfonts}
\usepackage{newtxtext,newtxmath}
\usepackage{hyperref}
\usepackage[ruled,linesnumbered,vlined]{algorithm2e}
\usepackage{graphicx}
\usepackage{textcomp}
\usepackage[dvipsnames, rgb]{xcolor}
\usepackage{balance}
\usepackage{verbatim}
\let\labelindent\relax
\usepackage{enumitem}
\usepackage{soul}
\usepackage{tablefootnote}
\usepackage{tikz}
\usetikzlibrary{shapes}
\makeatletter
\newcommand{\reducedplus}{\mathpalette\reduced@plus\relax}
\newcommand{\reduced@plus}[2]{%
  \sbox6{$\m@th#1+$}%
  \sbox8{\scalebox{0.875}{\copy6}}%
  \dimen@=\dimexpr(\wd6-\wd8)/3\relax
  \raisebox{\dimen@}{\box8}%
}
\newcommand{\boxoperation}[2][\mathbin]{%
  #1{\mathpalette\box@operation{#2}}%
}
\newcommand{\box@operation}[2]{%
  \ooalign{$\m@th#1\boxempty$\cr\hidewidth$\m@th#1#2$\hidewidth\cr}%
}

\def\secref#1{Sec.~\ref{#1}}
\def\figref#1{Fig.~\ref{#1}}
\def\tabref#1{Tab.~\ref{#1}}
\def\eqref#1{Eq.~(\ref{#1})}
\def\algref#1{Alg.~\ref{#1}}

\SetKwIF{Upon}{}{}{upon}{then}{}{}{}

\DeclareMathOperator{\argmin}{\underset{\textbf{x}}{\mathrm{argmin}}}

\tikzset{cross/.style={cross out, draw=black, minimum size=10pt, inner sep=0.5pt, outer sep=0.5pt}, cross/.default={5pt}}

\usepackage{fancyhdr}
\fancypagestyle{arxivhdr}
{
   \fancyhf{}
   \setlength{\headheight}{15pt} 
\fancyfoot[C]{This paper has been accepted for publication at the 2024 IEEE International Conference on Robotics and Automation (ICRA)DOI: 10.1109/ICRA57147.2024.10611214 \url{https://ieeexplore.ieee.org/document/10611214}}
\fancyhead[C]{\footnotesize Please cite this paper as:\\
E. Olivastri and A. Pretto, "IPC: Incremental Probabilistic Consensus-based Consistent Set Maximization for SLAM Backends," 2024 IEEE International Conference on Robotics and Automation (ICRA), 2024, pp. 10283-10289, doi: 10.1109/ICRA57147.2024.10611214}
}

\title{\LARGE \bf
IPC: Incremental Probabilistic Consensus-based Consistent Set Maximization for SLAM Backends
}
\author{Emilio Olivastri and Alberto Pretto% 
\thanks{
All the authors are with the Department of Information Engineering (DEI), University of Padova, Italy. {\tt\small Emails:[olivastrie; alberto.pretto]@dei.unipd.it}.
}}

\begin{document}

\newcommand{\bluecircle}{\raisebox{0pt}{\tikz{\node[draw,scale=0.5,circle,fill=black!20!blue](){};}}}
\newcommand{\pinktriangle}{\raisebox{0pt}{\tikz{\node[draw,scale=0.4,regular polygon, regular polygon sides=3,fill=pink!90,rotate=180](){};}}}
\newcommand{\greentriangle}{\raisebox{0pt}{\tikz{\node[draw,scale=0.4,regular polygon, regular polygon sides=3,fill=OliveGreen!,rotate=0](){};}}}
\newcommand{\cyansquare}{\raisebox{0pt}{\tikz{\node[draw,scale=0.5,regular polygon, regular polygon sides=4,fill=SkyBlue!80](){};}}}
\newcommand{\purplediamond}{\raisebox{0pt}{\tikz{\node[draw,scale=0.5,diamond,fill=blue!60!orange](){};}}}
\newcommand{\goldpentagon}{\raisebox{0pt}{\tikz{\node[draw,scale=0.6,regular polygon, regular polygon sides=5,fill=YellowOrange!70!](){};}}}
\newcommand{\redx}{\raisebox{-2pt}{\tikz{\node[draw,scale=0.5,cross,rotate=0,draw=red, line width=1mm=5mm](){};}}}
\newcommand{\browncross}{\raisebox{-2pt}{\tikz{\node[draw,scale=0.5,cross,rotate=45,draw=RawSienna!70!, line width=1mm=5mm](){};}}}

\maketitle
% Arxiv header
\thispagestyle{empty}
\pagestyle{empty}
\thispagestyle{arxivhdr}% Title page has fancy
%%%%%%%%%%%%%%%%%%%%%%%%%%%%%%%%%%%%%%%%%%%%%%%%%%%%%%%%%%%%%%%%%%%%%%%%%%%%%%%%
\begin{abstract}
In SLAM (Simultaneous localization and mapping) problems, Pose Graph Optimization (PGO) is a technique to refine an initial estimate of a set of poses (positions and orientations) from a set of pairwise relative measurements. The optimization procedure can be negatively affected even by a single outlier measurement, with possible catastrophic and meaningless results. Although recent works on robust optimization aim to mitigate the presence of outlier measurements, robust solutions capable of handling large numbers of outliers are yet to come.
This paper presents IPC, acronym for Incremental Probabilistic Consensus, a method that approximates the solution to the combinatorial problem of finding the maximally consistent set of measurements in an incremental fashion.
It evaluates the consistency of each loop closure measurement through a consensus-based procedure, possibly applied to a subset of the global problem, where all previously integrated inlier measurements have veto power.
We evaluated IPC on standard benchmarks against several state-of-the-art methods. 
Although it is simple and relatively easy to implement, IPC competes with or outperforms
the other tested methods in handling outliers while providing online performances.
We release with this paper an open-source implementation of the proposed method.
%This paper presents a method that approximates the solution to the combinatorial problem of finding the maximally consistent set of measurements in an incremental fashion. 
\end{abstract}
%State-of-the-art SLAM frameworks are split into two main modules: a front-end that processes raw data from the sensors and produces measurements,
%and a back-end that will use these measurements to estimate the system's state.
\section{Introduction}
%Solving the SLAM problem is a fundamental task in mobile robotic applications. 
A reliable, robust, and scalable solution to the SLAM problem is universally recognized as an enabling technology to make a robot truly autonomous. Although it is a problem that has been widely addressed by the research community and industry for more than 30 years, it cannot yet be considered a solved problem. For example, the design of fail-safe, self-tuning SLAM systems is still an open challenge, with many aspects still unexplored \cite{Cadena16tro-SLAMfuture}.
In recent years, the research community converged on modeling the SLAM problem through a graphical representation, in which nodes represent the poses of the robot, while the edges encode the probabilistic spatial relations between nodes. These constraints are computed from estimating the robot's motion, or by observing the environment and are affected by some degree of noise. Pose Graph Optimization (PGO) leverages the graphical representation employed in SLAM to frame and solve the non-convex maximum likelihood estimation (MAP) problem. The solution corresponds to the nodes' configuration that maximizes consistency between the constraints that define the problem. Libraries, such as g2o \cite{grisetti2011g2o}, GTSAM \cite{gtsam}, and Ceres \cite{Agarwal_Ceres_Solver_2022}, are commonly used to solve the PGO problem. \\ 
To build a pose graph, it is required a \textit{front-end} module that generates the spatial constraints. 
\begin{figure}[t!]
    \centering
    \begin{minipage}[t]{.23\textwidth}
        \centering
        \includegraphics[width=\textwidth]{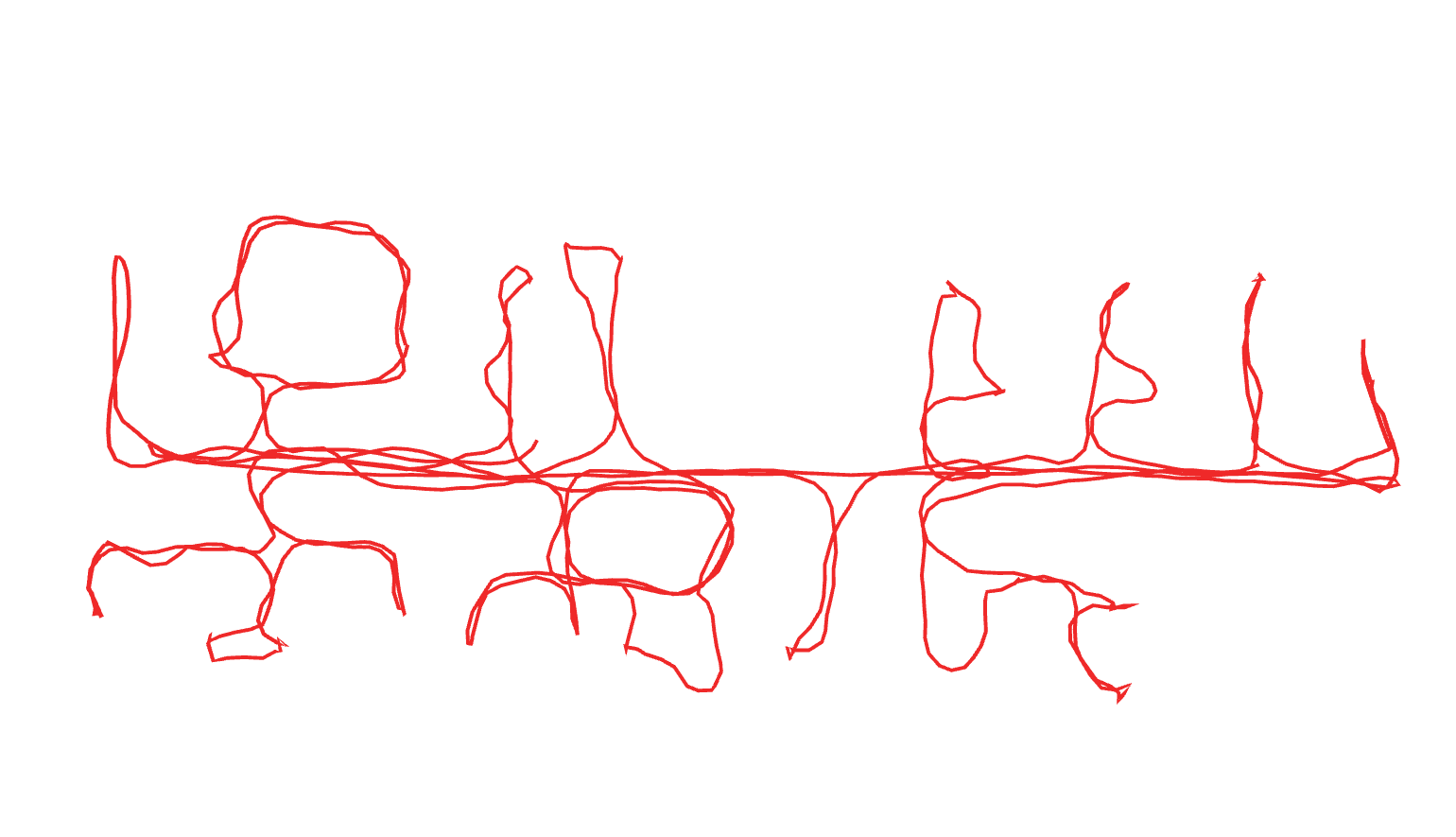}
        \center{\vspace*{-2ex}\footnotesize(a) Ground truth trajectory}
    \end{minipage}
    \begin{minipage}[t]{.23\textwidth}
        \centering
        \includegraphics[width=\textwidth]{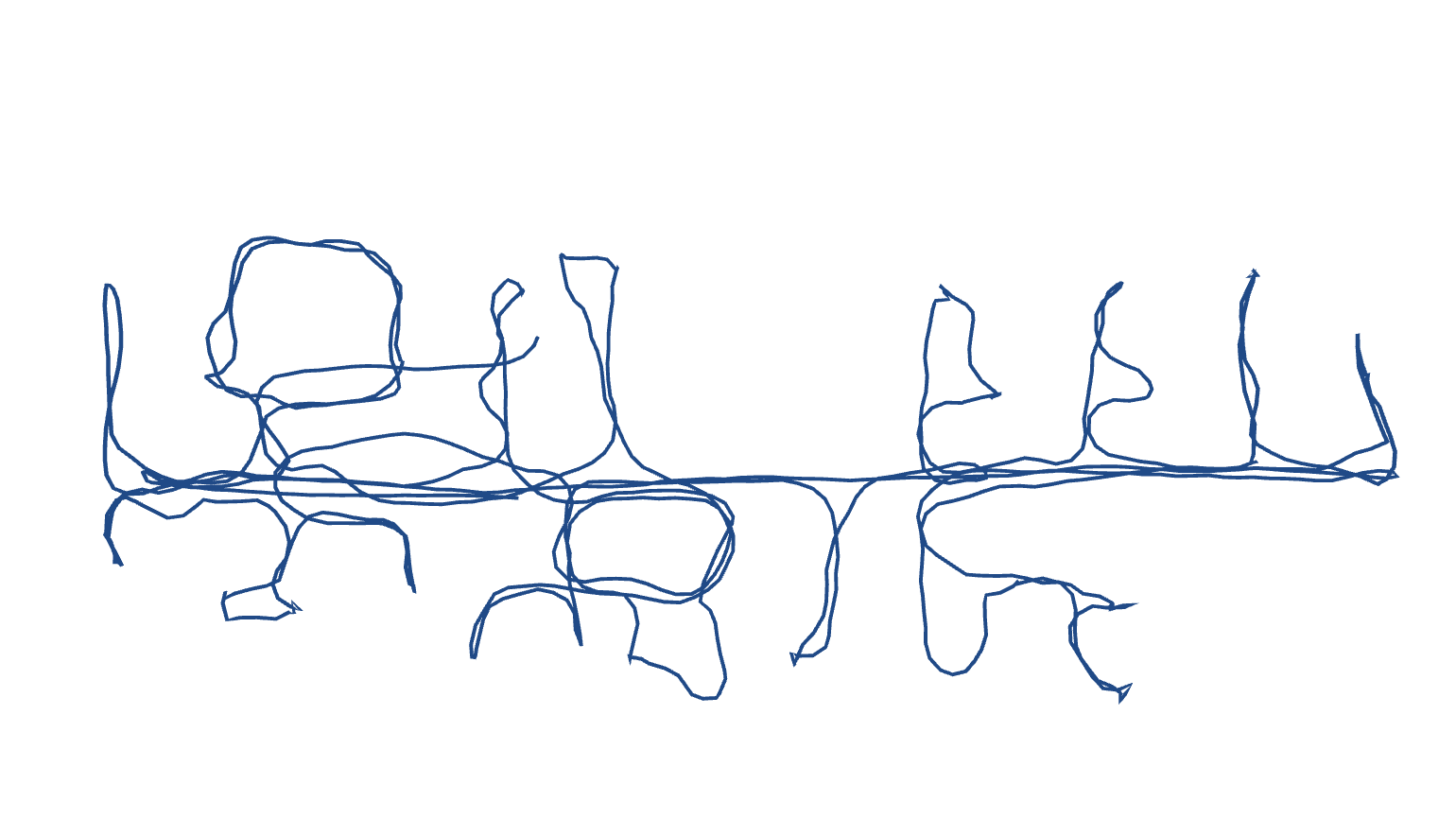}
        \center{\vspace*{-2ex}\footnotesize(b) IPC}
    \end{minipage}
    \begin{minipage}[t]{.23\textwidth}
        \centering
        \includegraphics[width=\textwidth]{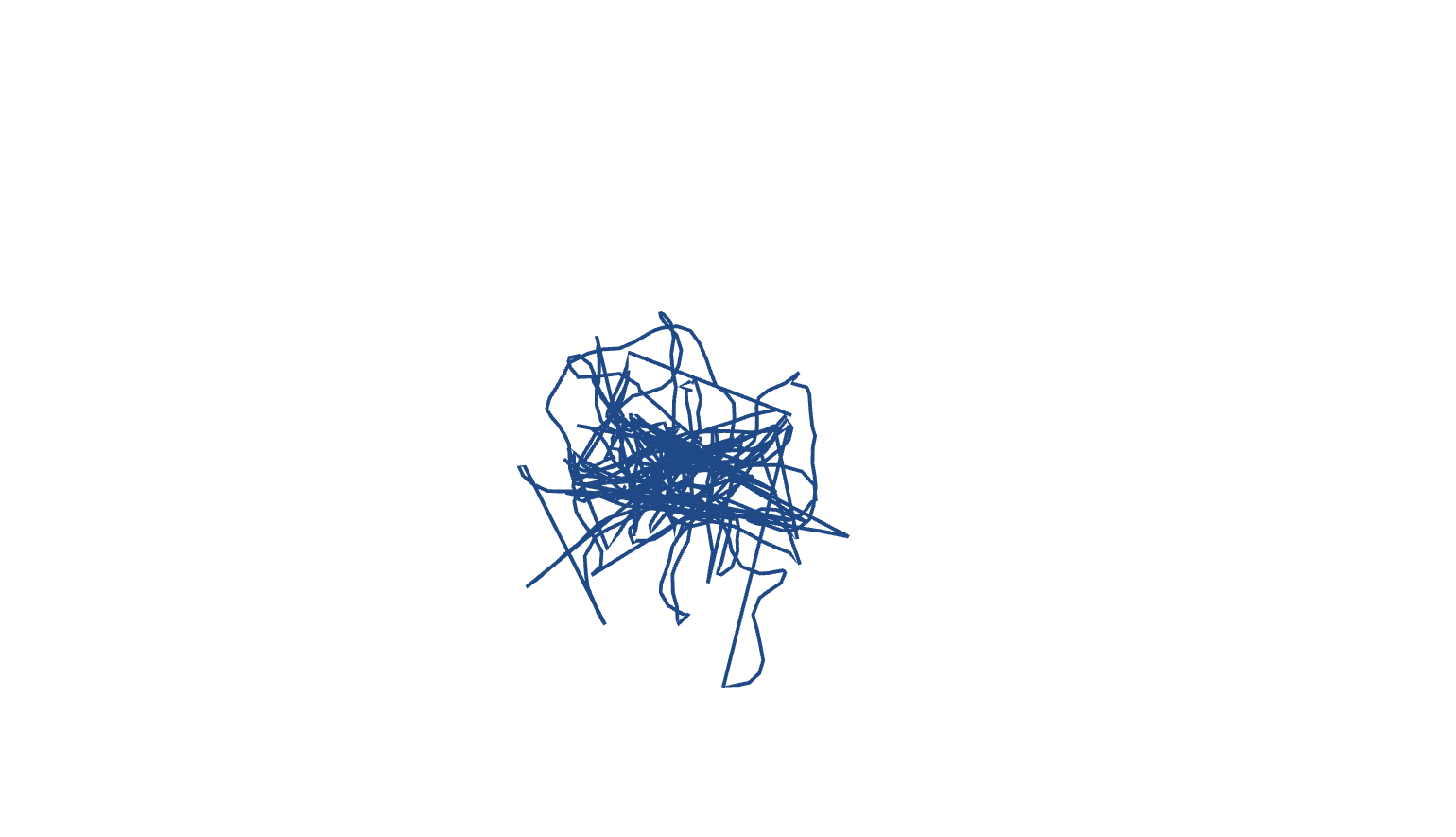}
        \center{\vspace*{-2ex}\footnotesize(c) ADAPT\cite{barron2019general}}
    \end{minipage}
    \begin{minipage}[t]{.23\textwidth}
        \centering
        \includegraphics[width=\textwidth]{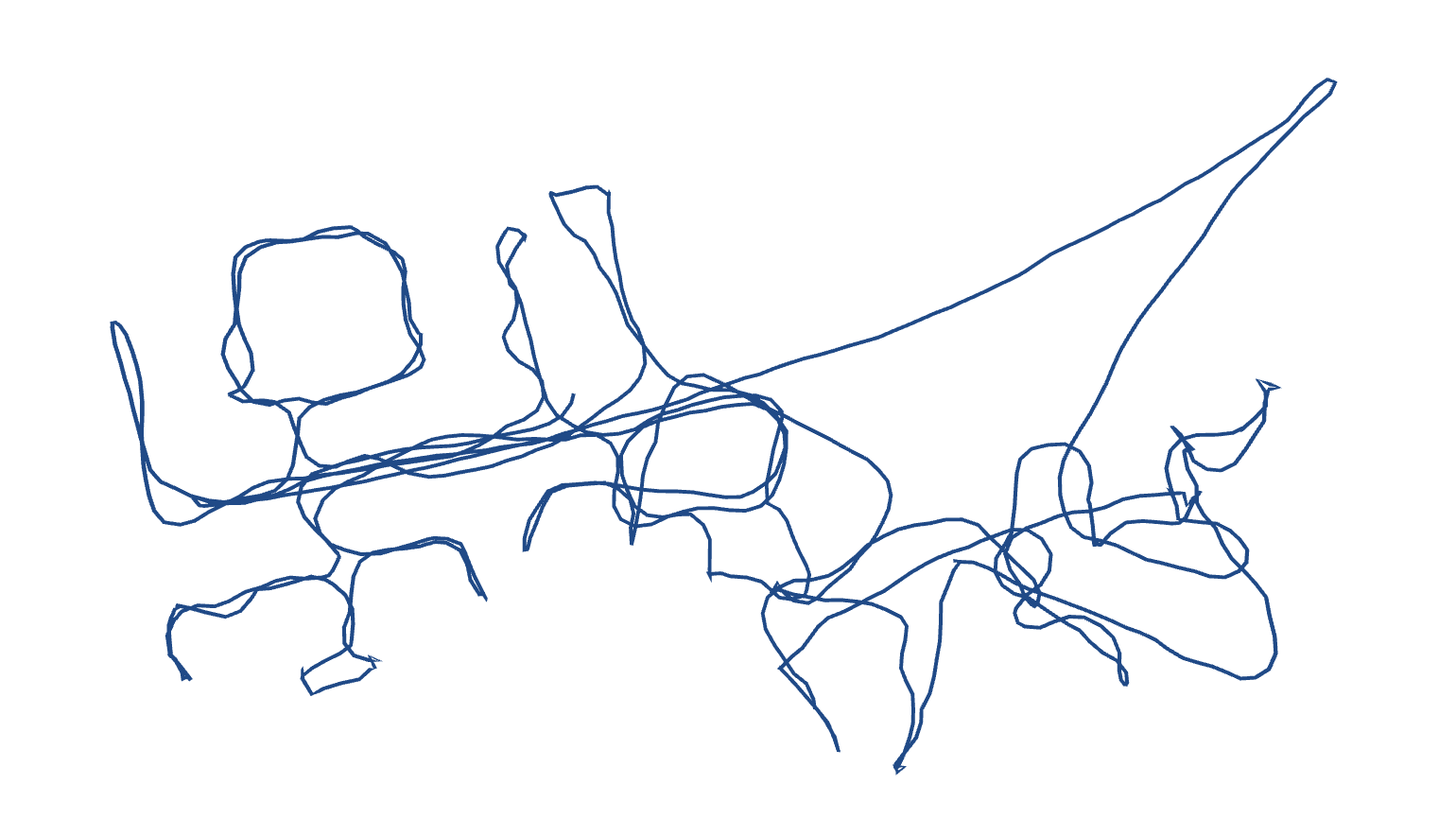}
        \center{\vspace*{-2ex}\footnotesize(d) DCS\cite{agarwal2013robust}}
    \end{minipage}
    \begin{minipage}[t]{.23\textwidth}
        \centering
        \includegraphics[width=\textwidth]{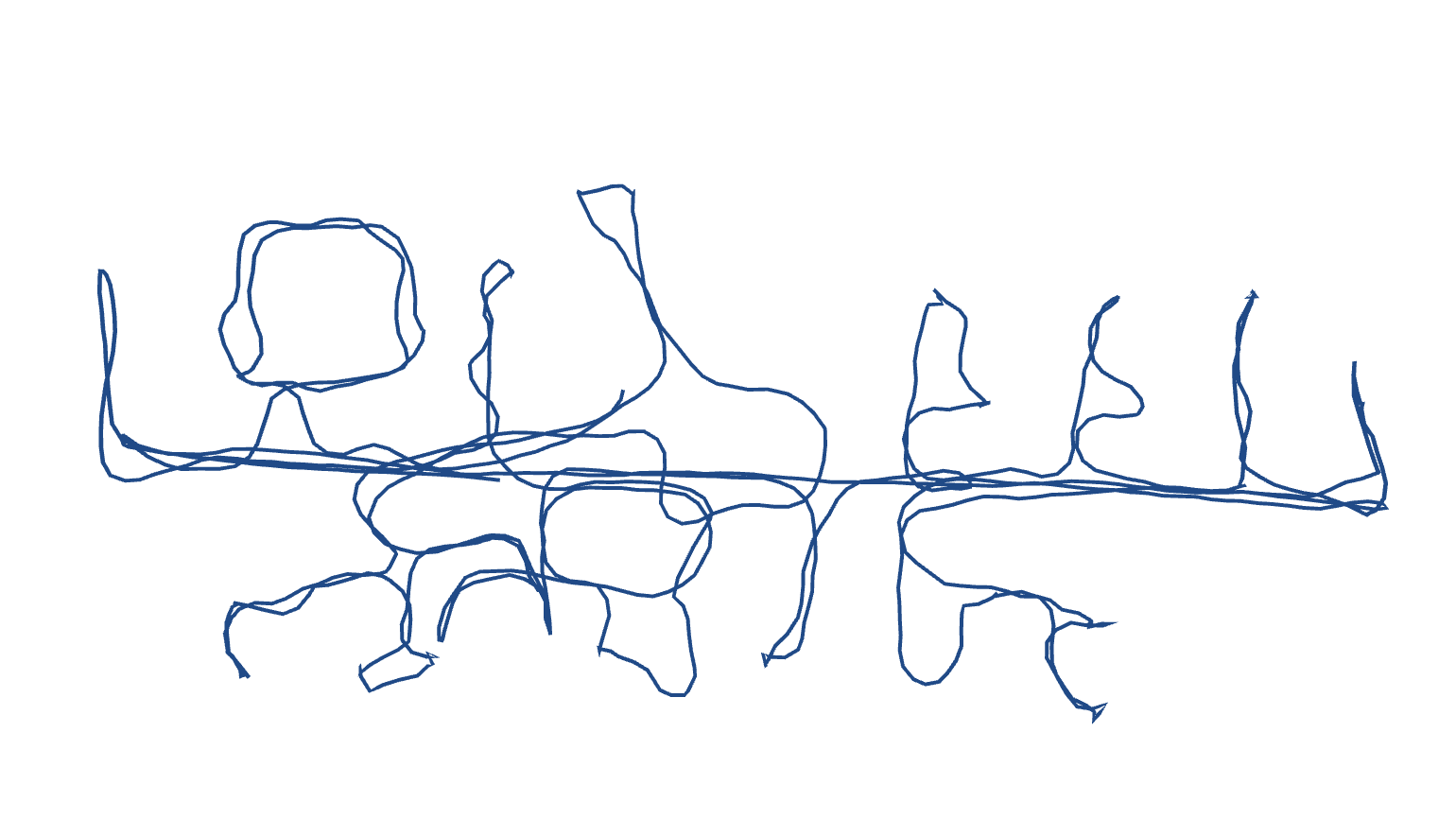}
        \center{\vspace*{-2ex}\footnotesize(e) GM\cite{black1996unification}}
    \end{minipage}
    \begin{minipage}[t]{.23\textwidth}
        \centering
        \includegraphics[width=\textwidth]{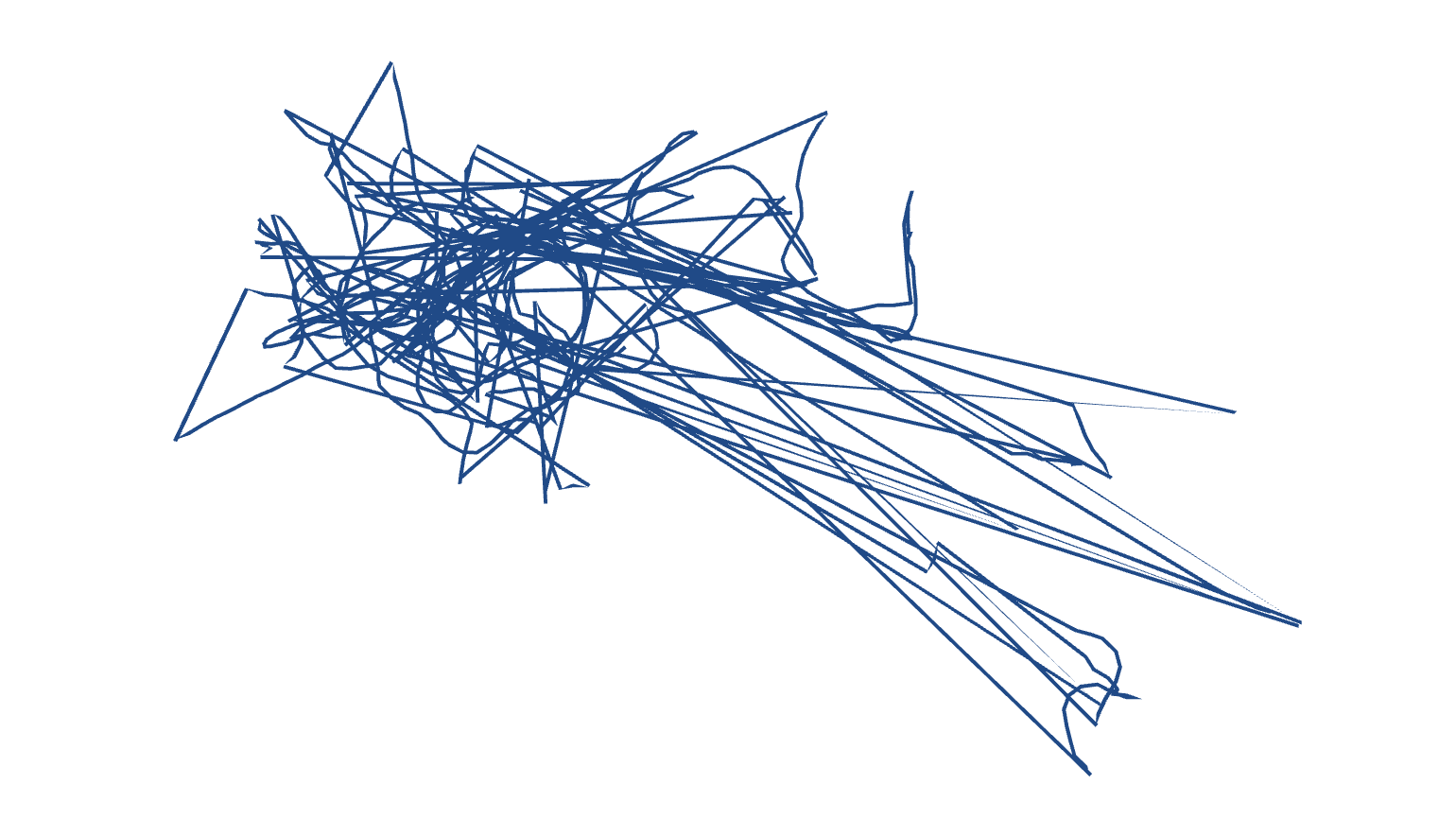}
        \center{\vspace*{-2ex}\footnotesize(f) GNC\cite{yang2020graduated}}
    \end{minipage}
    \begin{minipage}[t]{.23\textwidth}
        \centering
        \includegraphics[width=\textwidth]{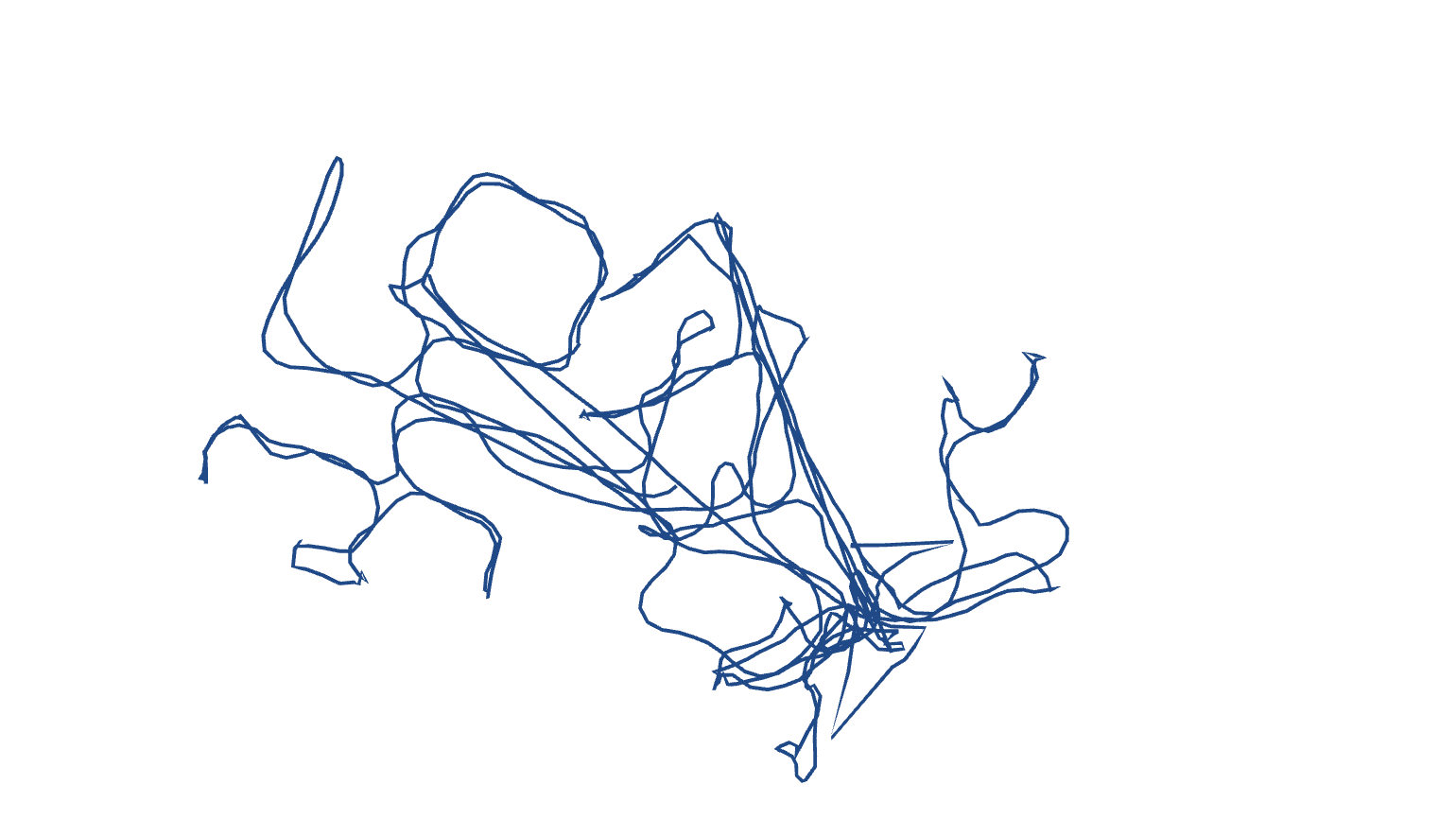}
        \center{\vspace*{-2ex}\footnotesize(g) PCM\cite{mangelson2018pairwise}}
    \end{minipage}
    \begin{minipage}[t]{.23\textwidth}
        \centering
        \includegraphics[width=\textwidth]{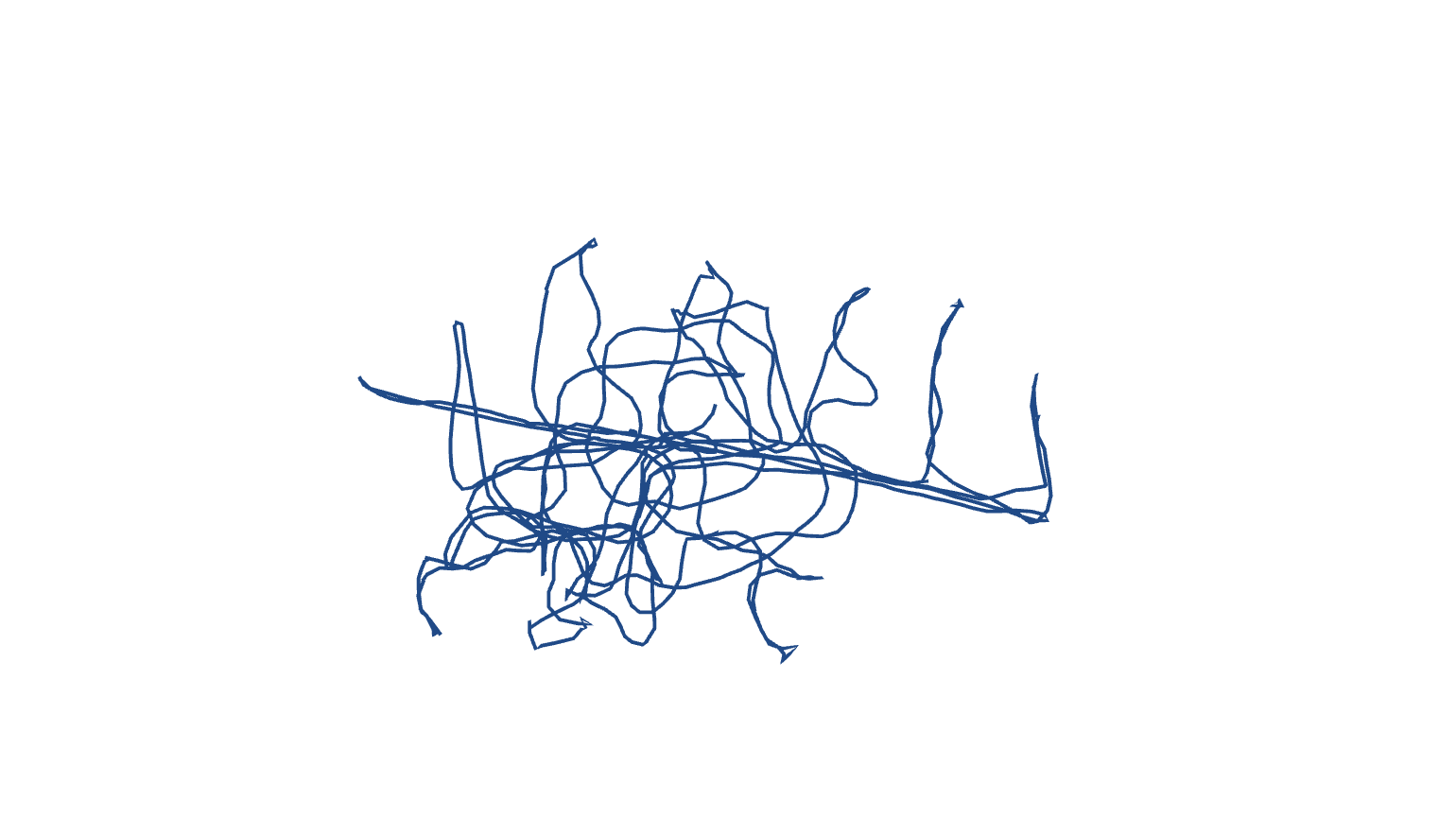}
        \center{\vspace*{-2ex}\footnotesize(h) MAXMIX\cite{olson2013inference}}
    \end{minipage}
    \caption{Trajectory estimated by the different methods on the FR079 dataset corrupted with 70\% of outliers.}
    \label{fig:intro_image}
    \vspace{-8mm}
\end{figure}
In this work, we are going to focus on two primary types of constraints: odometry constraints, which relate together consecutive poses, and loop closure constraints, which relate together non-consecutive poses. Most approaches assume these constraints are affected by noise but do not contain outliers.
Due to the inherent perceptual aliasing in the environment, these assumptions hold only when exploiting very conservative loop closure detectors, aiming at maximizing the probability of integrating a few inlier measurements at the cost of discarding, along with all outliers, most of the inliers. These behaviors are necessary because, in standard optimization methods, even a single outlier can completely distort the final results.
Nevertheless, it would be unrealistic to expect the front-end to never make incorrect assumptions based on isolated observations, given the environment's aliasing effects. Therefore, it should be the back-end's responsibility to handle erroneous measurements, due to its ability in accessing the full exploration history. \\
Recent works on graph-based SLAM proposed stacking methods that are able to handle large amounts of outliers, such as RANSAC\cite{matas2004randomized}, GNC\cite{yang2020graduated}, Consistent Set Maximization\cite{mangelson2018pairwise}\cite{lusk2021clipper}\cite{neira2001data}, and M-estimators\cite{barron2019general}, on top of classic solvers.
Our contribution falls in the Consistent Set Maximization category. The proposed approach tries to approximate the solution to the combinatorial problem of finding the maximally consistent set of measurements in an incremental fashion. While the majority of solutions are designed for offline usage, our approach is able to achieve online performances, making it suitable for use during missions rather than solely in post-processing.
For each incoming loop closure measurement, we find and optimize only a subgraph for which this constraint has an active influence. The solution to this subproblem is tested by using previously accepted measurements, each one with veto power, to decide whether to accept the new measurement. If accepted, also the subgraph optimization is propagated to the rest of the graph.\\
The main contributions of this work are the following:
\begin{enumerate}[label=(\roman*)]
%    \item A deep analysis of the typical assumptions taken into consideration when addressing Consensus-Maximization;
    \item IPC: An online, consensus-based approach able to incrementally build a good approximation of the maximally consistent set of loop closure measurements;
    \item An exhaustive evaluation of the proposed method. The evaluation is performed in standard datasets used for benchmarking state-of-the-art outlier rejection solutions;
    \item An open-source implementation of IPC available with this paper at:\\
\url{https://github.com/EmilioOlivastri/IPC}
    \normalsize
\end{enumerate}
The remainder of this paper is organized as follows:\\ \secref{sec:rel_works} reviews related work on robust optimization and outliers rejection. In \secref{sec:methodology} IPC and the building blocks from which it was defined are presented. In \secref{sec:experiments} we show a comprehensive evaluation of the current method and we compare it against state-of-the-art solutions on commonly used datasets. Finally, in \secref{sec:conclusions} we draw our conclusions.

\section{Related Work}
\label{sec:rel_works}
It is sufficient to have small percentages of outliers in the optimization process to invalidate the results completely. Robust pose graph optimization tries to mitigate the effect of such elements on the estimation process. M-estimators \cite{black1996unification} are one of the most classic techniques used in robust estimation. It involves the application of robust cost functions, such as Huber, Cauchy loss, German-McCloure, and Truncated-Least-Squares, to reduce the weight of the outliers outside of a "trust region", that defines the maximum accepted error. Once, the M-estimator is chosen, it is needed to find an effective threshold for the trusted area, which is a non-trivial task.
In order to bypass the problem of selecting the most suitable robust cost function, in \cite{barron2019general} an adaptive robust loss function (ADAPT) has been presented. Such loss is a generalization that is able to represent all the different robust cost functions using a control parameter. It is able to automatically determine which one is the most suitable for a particular instance of the problem. Another limitation that these approaches share is the introduction of extra non-linearity, which will likely introduce more local minima to the solution space.\\
To mitigate the impact of additional non-convexity, approaches such as Graduated Non-Convexity (GNC) \cite{yang2020graduated} have been proposed. This method involves a stepwise process of progressively introducing non-convexity to the convex approximation of the initial problem, ultimately returning it to its original formulation. When combined with TLS, GNC enables the optimizer to filter out outliers more efficiently compared to a direct application of the M-Estimator. The major drawback is the increase in convergence time. In \cite{mcgann2023robust}, thanks to the definition of the Scale Invariant Graduated(SIG) Kernel, the authors were able to adapt the GNC paradigm to an incremental solver such as iSAM2 \cite{kaess2012isam2}, reducing dramatically the number of iterations needed for convergence.
Another widely adopted approach is to let the optimizer directly handle the outliers detection by solving an augmented version of the original problem. For example, in \cite{sunderhauf2012switchable} latent variables that are able to activate and deactivate measurements are added to the original formulation of the problem. In \cite{agarwal2013robust}, the authors suggest an equivalent formulation that removes the additional "switchable" variables, reducing the convergence time of the method with respect to \cite{sunderhauf2012switchable}. In \cite{olson2013inference} the presence of outliers was modeled using a multi-modal Gaussian distribution. In particular, the authors proposed a max-mixture distribution to enable the optimizer to select the most promising Gaussian at each iteration.\\
% Qua invece parlare di maximum consensus set
Another line of research, under the assumption that inlier measurements are jointly consistent, tries to find the set of maximum size of consistent measurements instead of explicitly trying to detect outliers. Computing the maximally consistent set poses a challenge due to its combinatorial complexity, driving the research community to explore effective approximations for its resolution. Joint Compatibility Branch and Bound (JCBB) \cite{neira2001data} falls under the category of exact methods that are able to do better than exponential time, but are not viable for either real-time or online applications. In \cite{latif2013robust} the set of measurements is temporally divided into clusters, that are tested for both intra-consistency and inter-consistency. They approximate the maximum consensus set to all those measurements that passed both tests. In \cite{wu2020cluster} the consistency check was improved by adding a term that was able to mitigate the effect of noisy odometry measurements at the cost of a higher convergence time.\\
% Parlare del maximum consensus set con PCM/ROBIN e CLIPPER
Pairwise Consistency Maximization (PCM) \cite{mangelson2018pairwise} approximates joint consistency with pairwise consistency and by using approximate Max-Clique algorithms to compute the maximum pairwise consistent set. In ROBIN \cite{shi2021robin} the more general concept of invariance is introduced. It is used to define a new compatibility test from which to build the compatibility graph. Still, to establish effective invariants, it is essential to exploit priors that are inherent to the specific problem being addressed. In Clipper \cite{lusk2021clipper} the authors focused on defining different geometric-based invariants for solving diverse registration problems. The method is able to detect the densest compatibility graph while maintaining a relatively low execution time through the projection gradient ascent with a back-tracking line search.  

\section{Methodology}
\label{sec:methodology}
We expect the robot to be equipped with (i) an ego-motion estimation front-end (e.g.,  wheel odometry or visual-inertial odometry) that provides over time the rigid transformations between consecutive poses $\mathbf{x}_a, \mathbf{x}_{a+1} \in SE(2)~\text{( or}~SE(3)\text{)}$ traversed by the robot; (ii) a place recognition module that can detect medium- and long-term previously seen locations and estimate the transformation between the related previous pose $\mathbf{x}_a$ and the current one $\mathbf{x}_b$ (i.e., a \emph{loop closure measurement} or \emph{constraint}). Given the sequence of odometry and loop closure measurements, our method aims to find the maximum consistent set of measurements while estimating the whole trajectory followed by the robot.
\subsection{Assumptions and Heuristics}
\label{sec:assumption}
Our method builds upon the following assumptions:\\
\textbf{A1}:\textit{Odometry measurements are inliers.} The quality of the odometry can change based on the front-end that is used, but as stated in \cite{carlone2014selecting}, they are usually not corrupted by outliers. These measurements are reliable constraints that can be used to enforce maintaining the local shape of the trajectory.\\
%Consequently, the task of assessing consistency between constraints is primarily focused on loop closure edges.\\ <-- Non serve
\textbf{A2}:\textit{ Prevalence of inliers.} 
Given the inherent aliasing effect in the environment, it is normal for the place recognition module to produce a certain percentage of outliers. On the other hand, it is reasonable to expect that the majority of loop closure measurements are inliers.
Furthermore considering \textbf{A1} and the fact that the frequency of odometry measurements is generally higher than the one of loop closure measurements, the prevalence of inlier measurements is guaranteed.\\
As generally done in robust optimization, in IPC we apply the following heuristics:

\textbf{H1}:\textit{ Start from a simpler version of the problem, for which its solution will be meaningful for the more complex version of the problem.} As stated in \cite{grisetti2012robust} and later in \cite{guadagnino2021hipe}, a simpler version of a problem exhibits a broader and more stable convergence basin, thus ensuring better convergence. Consequently, the solution obtained for the reduced problem can be used as the initial guess for the full problem, also improving the convergence of the latter.\\ 
\textbf{H2}:\textit{ The nature of the compatibility test is to reject as many outliers as possible while preserving the inliers.} The distinction between outliers and inliers is not always observable as stated in \cite{carlone2014selecting}, so instead of trying to detect which measurements are outliers, consistency can be verified using \eqref{eq:consensus}.\\
%Here instead we are going to outline some of the main heuristics used in robust optimization that were applied also for IPC:\\ 
\subsection{Problem Formulation}
A pose graph $\mathcal{G} = (\mathcal{V},\mathcal{E})$ uses nodes in $\mathcal{V}$ to represent the discretized poses $\mathbf{x}_i$ of the robot (i.e., the followed trajectory) while the edges $(a,b) \in \mathcal{E}$ that connect the nodes represent the spatial relations (i.e., the measurements, or constraints) between them.
The full state can be represented using the vector $\mathbf{x}= [\mathbf{x}^\top_1, ..., \mathbf{x}^\top_n]^\top$, where the subscript defines the temporal order
 and $n$ is the total number of poses. The goal of PGO is to find a solution $\mathbf{x^*}$ that minimizes the following objective function:
\begingroup
\setlength\abovedisplayskip{4pt}
\begin{equation}
    \label{eq:full_problem}
    \mathbf{x^*} = \argmin\sum_{(a,b) \in \mathcal{E}} \mathbf{e}_{ab}(\mathbf{x}_a, \mathbf{x}_b)^T \Omega_{ab}  \mathbf{e}_{ab}(\mathbf{x}_a, \mathbf{x}_b)
\end{equation}
\endgroup
where $\Omega_{ab}$ is the information matrix associated with the measurement collected between $\mathbf{x}_a$ and $\mathbf{x}_b$ and the error is defined as follows : 
\begingroup
\setlength\abovedisplayskip{2pt}
\begin{equation}
    \label{eq:general_error}
   \mathbf{e}_{ab}(\mathbf{x}_a, \mathbf{x}_b) = \mathbf{h}(\mathbf{x}_a, \mathbf{x}_b) \boxminus \mathbf{z}_{ab}.
\end{equation}
\endgroup
where $\mathbf{h}(\mathbf{x}_a, \mathbf{x}_b)$ is a function that computes the rigid transformation between the two poses, $\mathbf{z}_{ab}$ is the odometry or loop closure measurement, and $\boxminus$ is the difference operator for the manifold domain. We use the notation $\mathbf{x}_i$ to both refer to a robot pose and the corresponding node, and $\mathbf{e}_{ab}$ to both refer to an error term and the corresponding edge of $\mathcal{G}$.
Following \cite{latif2013robust}, we partition the edges in two sets, $\mathcal{E} = \mathcal{E}_o \cup \mathcal{E}_l $, where $\mathcal{E}_o = \{ (a,b) \in \mathcal{E} \;|\; b = a + 1\}$ is the set of edges associated with the odometry measurements, and $\mathcal{E}_l = \{ (a,b) \in \mathcal{E} \;|\; b \neq a+ 1\}$ is the set of edges 
 associated with the loop closure measurements.\\ To make the notation more fluid the following shortcuts are used: 
\begin{itemize}
    \item $\mathbf{e}_a = \mathbf{e}_{a, a+1}(\mathbf{x}_a, \mathbf{x}_{a+1})$ for the odometry errors;
    \item $ \mathbf{e}_{ab} = \mathbf{e}_{ab}(\mathbf{x}_a, \mathbf{x}_{b})$ for the loop closure.
\end{itemize} 
It allows us to rewrite \eqref{eq:full_problem} as:
\vspace{-2mm}
\begin{multline}
    \label{eq:offline_division1}
    \mathbf{x^*} = \argmin\sum_{(a,a+1) \in \mathcal{E}_o} \mathbf{e}_a^T \Omega_{a,a+1}  \mathbf{e}_a + \sum_{(a,b) \in \mathcal{E}_l} \mathbf{e}_{ab}^T \Omega_{ab}  \mathbf{e}_{ab}
\end{multline}
\vspace{-5mm}
\begin{figure}[t!]
    \centering
    \includegraphics[width=0.5\textwidth]{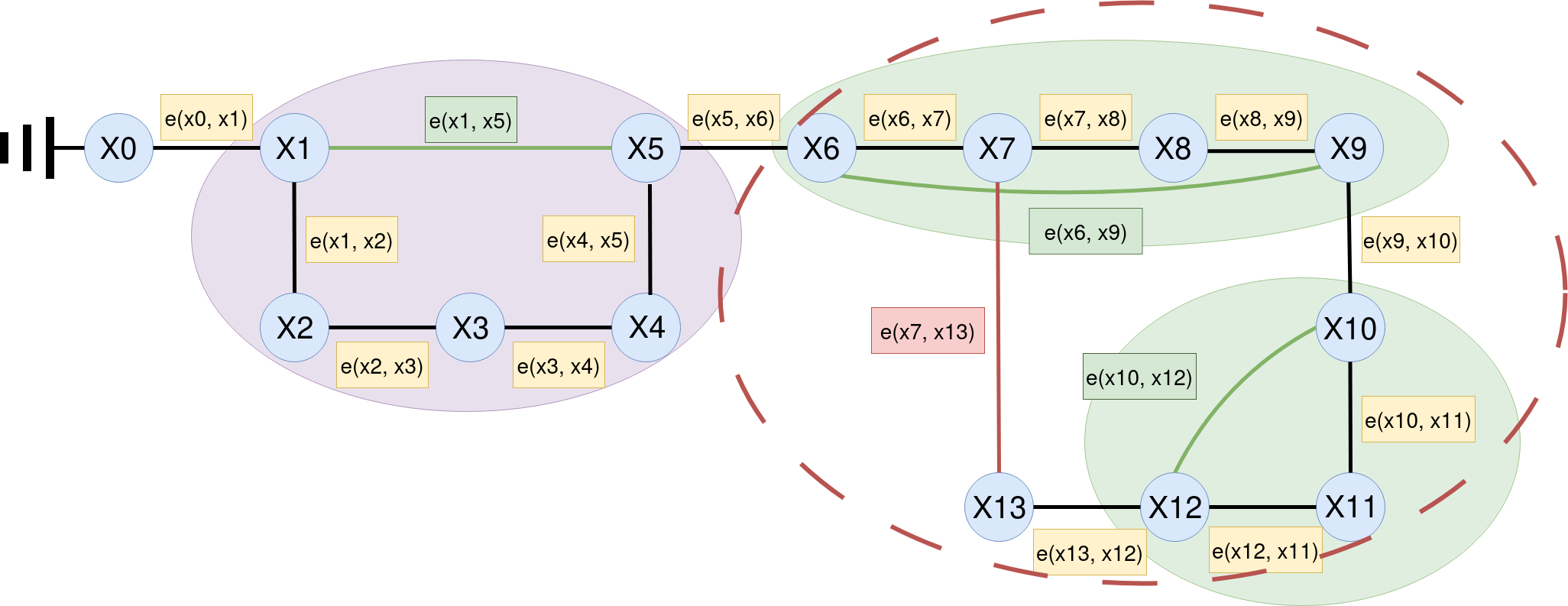}\hfill
    \caption{An example of pose graph with nodes represented by light blue circles. Each error term is associated with the corresponding edge. The coordinate system is usually fixed in the first node $\mathbf{x}_0$. Light green and purple ovals represent examples of simple independent subgraphs (i.e., they only include one loop) while the dotted red oval represents more complex independent subgraphs, that include both multiple loops with crossing edges and internal loops.}
    \label{fig:strict_consensus}
    \vspace{-5mm}
\end{figure}
\subsection{Independent Subgraphs and Solution Propagation}
\label{sec:issp}
In the simplest case, a loop closure measurement (e.g., the green edge between $\mathbf{x}_1,$ and $\mathbf{x}_5$ in \figref{fig:strict_consensus}) creates in the pose graph $\mathcal{G}$ a loop without edge crossing or internal loops; all nodes that compose this loop form a subgraph $\mathcal{G}'$ (in the example, $\mathcal{G}'$ is composed of the set $\{\mathbf{x}_1,\dots,\mathbf{x}_5\}$, highlighted in purple) that can be optimized independently from the other nodes in $\mathcal{G} \setminus \mathcal{G}'$.
A subgraph defined in this way always defines a time interval that includes all nodes, and related edges, from index $a$ to index $b$, with $a<b~\text{and}~b-a > 1$. 
More formally, we can define $\mathcal{G}' = \mathcal{G}(a,b) = (\mathcal{V}(a,b),\mathcal{E}(a,b))$ where $\mathcal{V}(a,b)=\{\mathbf{x}_i|a \leq i \leq b\}$ and $\mathcal{E}(a,b)=\{\mathbf{e}_{ij}|a \leq i,j \leq b\}$. If no other node $\mathbf{x}_k  \in \mathcal{G} \setminus \mathcal{G}(a,b)$ is able to generate a loop with any of $\mathbf{x}_i \in \mathcal{G}(a,b)$, we define $\mathcal{G}(a,b)$ as an \emph{independent subgraph} $\mathcal{G}^I(a,b)$ that can be optimized independently from all the other nodes (see \figref{fig:strict_consensus}). The result of this optimization can be \emph{propagated} to all other nodes $\mathbf{x}_k \in \mathcal{G} \setminus \mathcal{G}'$ simply by applying the estimated transformation updates of the node in $\mathcal{G}'$ they are connected to. 
%For example, in \figref{fig:strict_consensus}) the subgraph $\mathcal{G}^I(1,5)$ that is composed of the nodes $\{\mathbf{x}_1,\dots,\mathbf{x}_5\}$ can be optimized independently of all other nodes; the result of this optimization can be propagated to the nodes $\mathbf{x}_k \in \mathcal{G}(6,13)$ by adding the transformation update applied to $\mathbf{x}_5$ thanks to the optimization.\\
For example, after optimizing the subgraph highlighted in purple in \figref{fig:strict_consensus}, the result of this optimization can be propagated to the nodes $\mathbf{x}_k \in \mathcal{G}(6,13)$ by adding the transformation update applied to $\mathbf{x}_5$.\\
Independent subgraphs can contain multiple loops and cross edges: for example in \figref{fig:strict_consensus} the arrival of a loop closure between node 7 and node 13 causes the fusion between the two previous independent subgraphs represented by the light green ovals.  
%We then define the set of variables \textbf{affected} by the loop closure constraint $e_{ab}$ as $X_{aff}^{ab} = \{ \mathbf{x}_c \;|\; c \in [a, b]\}$.
\subsection{The IPC Algorithm}
\begin{algorithm}[t!]
\DontPrintSemicolon
\caption{IPC}\label{alg:cap}
\small
$CnS, \mathcal{E}_o, \mathcal{E}_l, C \leftarrow \emptyset$\;
$\mathcal{V} \leftarrow \{\mathbf{x}_0\}$\;
\Upon{receiving $\mathbf{z}_{ij}$}
{
    \tcp{odometry measurement}
    \If{ $j = i + 1$  }
    {\label{alg:odometry}
        $\mathbf{x}_{i+1} \leftarrow \mathbf{x}_i \boxplus \mathbf{z}_{i,i+1}$ \tcp{integrate motion} \label{alg:motion}
        $\mathbf{e}_i \leftarrow \texttt{edge}(\mathbf{x}_{i}, \mathbf{x}_{i+1}, \mathbf{z}_{i,i+1}, \Omega_{i,i+1})$\;
        $\mathcal{E}_o \leftarrow \mathcal{E}_o \cup \{\mathbf{e}_i\}$ \tcp{update edges} \label{alg:update_Eo}
        $\mathcal{V} \leftarrow \mathcal{V} \cup \{\mathbf{x}_{i+1}\}$ \tcp{update variables}
    }
    \tcp{loop closure measurement}
    \Else
    {   \label{alg:loop}
        \tcp{find initial $\mathcal{G}^I$}
        $\mathbf{e}_{ij} \leftarrow \texttt{edge}(\mathbf{x}_{i}, \mathbf{x}_{j}, \mathbf{z}_{ij}, \Omega_{ij}) $\;
        $\mathcal{V}^I, \mathcal{E}^I \leftarrow find$  $\mathcal{G}^I_{ij}$\; \label{alg:initAff}
        \tcp{find cross edges until independent}
        $ a \leftarrow i $ \;
        \While{$\exists \mathbf{x}_k \in \mathcal{G} \setminus \mathcal{G}^I |\; \exists \mathbf{e}_{kg} \in CnS, a < g < j$}
        {
            $\mathcal{E}^I \leftarrow \mathcal{E}^I \cup \{\mathbf{e}_k,\dots, \mathbf{e}_a\}$\;
            $\mathcal{E}^I \leftarrow \mathcal{E}^I \cup \{\mathbf{e}_{kg}\}$\;
            $\mathcal{V}^I \leftarrow \mathcal{V}^I \cup \{\mathbf{x}_k,\dots, \mathbf{x}_{a-1}\}$\;
            $ a \leftarrow k$
        }\label{alg:end_aff}
        $\mathcal{E}_{so}, \mathcal{E}_{sl} \leftarrow \mathcal{E}^I$ \;
        \texttt{solve} \eqref{eq:complex_consensus} \tcp{PGO} \label{alg:opt}
        \If{ $\mathbf{e}_{ab}^T \Omega_{ab} \mathbf{e}_{ab} < \chi^2_{\alpha, \delta}, \forall e_{ab} \in \mathcal{E}^I$ } 
        {\label{alg:consensus}
            $CnS \leftarrow CnS \cup \{\mathbf{e}_{ij}\}$\; \label{alg:consSet}
            $\mathcal{E}_l \leftarrow \mathcal{E}_l \cup \{\mathbf{e}_{ij}\}$\;
            propagate solution to $\mathbf{x}_k \in \mathcal{V} \setminus \mathcal{V}^I$ as described in \secref{sec:issp} \label{alg:propagate}
        }
        \Else
        {
            $\texttt{restore}(\mathcal{V}^I)$\; \label{alg:restore}
        }
    }
}
\end{algorithm}
The IPC algorithm is fundamentally very simple and exploits and updates a set $CnS$ (called \textbf{consensus set}) of loop closure measurements that have been accepted in previous iterations. The following steps briefly outline the IPC algorithm, where we report in square brackets the used assumptions and heuristics (see \secref{sec:assumption}):
\begin{itemize}
    \item Integrates all odometry measurements and generates a new node when necessary (for example, if the distance to the previous node is above a certain threshold)[\textbf{A1}];
    \item For each incoming loop closure measurement $\textbf{z}'$, find the (minimal) independent subgraph $\mathcal{G'}$ (see \secref{sec:issp}) that contains such constraint by considering odometry constraints and only the loop closure constraint in $CnS$ collected so far;
    \item Solve the PGO problem only for the independent subgraph $\mathcal{G'}$ [\textbf{H1}];
    \item Accept and include $\textbf{z}'$ in $CnS$ only if \emph{all} measurements in $\mathcal{G'}$ are consistent (i.e., \emph{agree}, see below) with the new solution obtained in the previous point [\textbf{A1},\textbf{A2}, \textbf{H1},\textbf{H2}];
    \item If $\textbf{z}'$ has been accepted, propagate the obtained solution to the remaining nodes as described in \secref{sec:issp}.
\end{itemize}
To check whether a measurement agrees with a PGO solution, we exploit the $\chi^2$ test\cite{neira2001data}:
\begin{equation}
    \label{eq:consensus}
    \mathbf{e}_{ab}^T \Omega_{ab} \mathbf{e}_{ab} < \chi^2_{\alpha, \delta}, \forall e_{ab} \in \mathcal{E'}
\end{equation}
where $\alpha$ is the confidence parameter and $\delta$ the degrees of freedom of the error and $\mathcal{E'}$ is the edgset of $ \mathcal{G'}$. The error term is computed here by using as robot pose the solution obtained by the PGO carried out on $\mathcal{G'}$. If one of the tests fails, the new loop closure measurement $\textbf{z}'$ \emph{is not} accepted, hence not included in $CnS$.\\
We report the pseudocode of our method in \algref{alg:cap}, which is described step by step below. For the sake of simplicity, we assume here to create a new node for each new odometry measurement.\\
As soon as a new measurement $\mathbf{z}_{ij}$ is received by the system, the first operation to carry out is to identify if it is an odometry or a loop closure measurement. If an odometry measurement is received, then $ \mathbf{z}_{ij} = \mathbf{z}_{i,i+i}$ (line \ref{alg:odometry}), a new variable $\mathbf{x}_{i+1}$ is initialized integrating the motion estimate $\mathbf{z}_{i, i+1}$ from $\mathbf{x}_i$(line \ref{alg:motion}), and then added to $\mathcal{V}$. The final operation is to update the set $\mathcal{E}_o$ with the new edge $\mathbf{e}_i$ relative to the measurement $\mathbf{z}_{i, i+1}$(line \ref{alg:update_Eo}).\\
In the scenario a loop closure measurement $\mathbf{z}_{ij}$ is received (line \ref{alg:loop}), its relative edge $\mathbf{e}_{ij}$ is created. Then, $\mathbf{e}_{ij}$ is used to identify the minimal independent subgraph $\mathcal{G}^I(a, j)$ (line \ref{alg:initAff}-\ref{alg:end_aff}) such that :
\begin{equation}
    !\exists \mathbf{x}_k \in \mathcal{G} \setminus \mathcal{G}^I(a, j) |\; \exists \mathbf{e}_{kg} \in CnS, a < g < j
\end{equation}
Once, $\mathcal{G}^I(a, j)$ has been identified, its edges are decomposed in the following manner $ \mathcal{E} = \mathcal{E}_o \cup \mathcal{E}_l$. This step is fundamental because being the odometry measurements the "trustworthy" ones, a scaling factor $s > 1$ is applied to their information matrix to increase the importance of minimizing their error during the optimization using \eqref{eq:complex_consensus} (line \ref{alg:opt}):
\begin{equation}
        \label{eq:complex_consensus}
    \sum_{a \in \mathcal{E}_o} \mathbf{e}_a^T s\Omega_{a,a+1}  \mathbf{e}_a +
    \sum_{(a,b) \in \mathcal{E}_l} \mathbf{e}_{ab}^T \Omega_{ab}  \mathbf{e}_{ab}
\end{equation}
It allows the solver to prioritize maintaining the local shape of the trajectory, actively reducing the acceptance rate of outliers.
The scaling factor is fixed and doesn't play a role in the optimization procedure, thus avoiding an increase in the problem's complexity, as opposed to SC\cite{sunderhauf2012switchable} where additional extra variables were added. 
In case \eqref{eq:consensus} is not satisfied, it means that with the addition of $\mathbf{e}_{ij}$ the optimizer is not able to find a configuration of the nodes for which the new edge is consistent with the measurements in its $\mathcal{G}^I$ and the elements of $\mathcal{V}^I$ are restored to its state previous the optimization. Instead, if \eqref{eq:consensus} is satisfied, a consistent configuration is found leading to the acceptance of the loop closure to the $CnS$(line \ref{alg:consSet}). The current solution obtained for the nodes $\mathcal{V}^I$ is then propagated to the nodes that didn't participate in the optimization $\mathcal{V} \setminus \mathcal{V}^I$(line \ref{alg:propagate}).
\section{Experiments}
\label{sec:experiments}
We implemented the proposed method (called IPC in all experiments) in C++ building upon the g2o\cite{grisetti2011g2o}  framework, exploiting the Dog-Leg optimization steps and setting $s = 3$ for all experiments.
We compared IPC with 7 recent alternative, state-of-the-art approaches, reported below along with the acronym used in the plots and the used implementation. Most of these methods aim to mitigate the anomalous problem in loop closure detection.
For Max-Mixture method \cite{olson2013inference} (MAXMIX in the plots) in the experiment we used the implementation provided in OpenSLAM\footnote{\url{https://openslam-org.github.io/}}. We implemented inside g2o the Adaptive Kernel \cite{barron2019general} robust kernel (ADAPT in the plots). For Pairwise consistent measurement set maximization method \cite{mangelson2018pairwise} (PCM in the plots), Kimera's implementation is used\cite{rosinol2021kimera}.
The remaining methods are provided by the gtsam library\cite{gtsam}: Graduated Non-Convexity \cite{yang2020graduated}, Dynamic Covariance Scaling \cite{agarwal2013robust}, Huber and German-McClure \cite{black1996unification}, will be referred to in the plots respectively as GNC, DCS, HUBER, and GM. For all methods we kept all their parameters constant, reporting for each one of them the results obtained with the best set of parameters. Experiments were carried out in a cluster PC using a single Intel(R) Xeon(R) Gold 5220 CPU.
\subsection{Datasets}
The datasets used for evaluation are currently the standard for evaluating the performances of back-end optimizers and are described in \cite{carlone2014angular} \cite{carlone2014fast}. They can be divided into two categories, synthetic and real. The Intel, Mit, Csail, Freiburg Building (\figref{fig:intro_image}), and Freiburg University Hospital are real datasets, while M3500 is synthetic. In both cases, the datasets are outlier-free. 
The loop closure edges in the unaltered datasets were used as the ground truth for the inliers. Outliers were added, using the Vertigo package \cite{vertigo}, as a percentage of the total inliers present in the dataset in order to have balanced results.

\subsection{Metrics}
The \textbf{Precision}, \textbf{Recall}, and \textbf{F1 score}  metrics are used in the evaluation, the latter defined from the former:
\begin{equation*}
    Prec = \frac{TP}{TP + FP},  \; Rec = \frac{TP}{TP + FN},  \; F1 = 2 * \frac{Prec * Rec}{Prec + Rec}
\end{equation*}
where \textbf{TP}, \textbf{FP}, \textbf{TN} \textbf{FN} stand for a loop closure that is a true and false positive and true and false negative, respectively. Finally, the solution obtained using SE-Sync \cite{rosen2019se}, which is a fast non-minimal solver for PGO, on the outlier free version of the dataset was used as ground truth trajectory for the evaluation of both \textbf{Absolute Trajectory Error} (ATE) and \textbf{Relative Pose Error} (RPE)\cite{zhang2018tutorial}, that will be used to measure the accuracy of estimated trajectory. 

\begin{figure*}[t]
    \centering
    \includegraphics[width=0.9\linewidth]{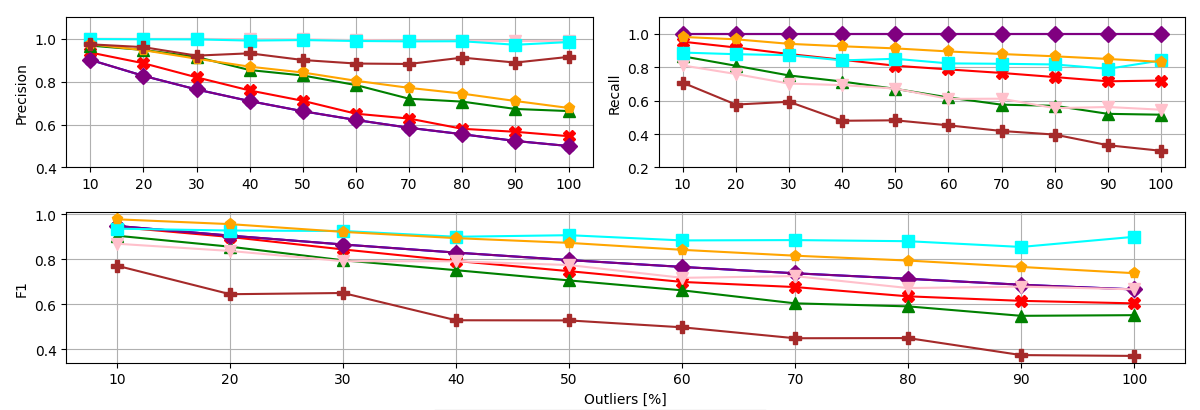}
    \vspace{-3mm}
    \caption{Top left, top right, and bottom images are respectively the precision, the recall, and F1 score performances of the methods for varying outlier percentages. Legend of the graphs: HUBER (\protect\greentriangle), GM(\protect\bluecircle), IPC(\protect\cyansquare), DCS(\protect\purplediamond), MAXMIX(\protect\pinktriangle), ADAPT(\protect\goldpentagon), PCM(\protect\browncross), GNC(\protect\redx). GM is completely covered by DCS in both precision and recall plots, while MAXMIX is nearly entirely covered by IPC in the precision plot.}
    \label{fig:full_stats}
\end{figure*}
\begin{figure*}[t]
\vspace{-3mm}
    \centering
    \includegraphics[width=0.9\linewidth]{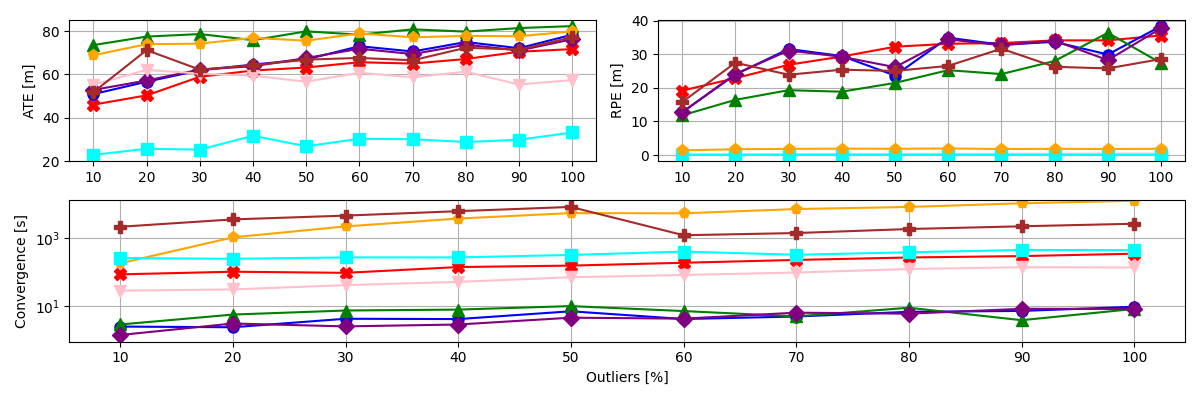}
    \vspace{-3mm}
    \caption{Top left, top right, and bottom images are respectively the ATE, the RPE, and convergence time of the methods for varying outlier percentages. Legend of the graphs: HUBER (\protect\greentriangle), GM(\protect\bluecircle), IPC(\protect\cyansquare), DCS(\protect\purplediamond), MAXMIX(\protect\pinktriangle), ADAPT(\protect\goldpentagon), PCM(\protect\browncross), GNC(\protect\redx). The MAXMIX plot for the RPE metrics is totally covered by the ADAPT plot.}
    \label{fig:ate_rpe}
    \vspace{-5mm}
\end{figure*}

\begin{table}[t!]
\centering
\resizebox{0.9\linewidth}{!}{
\begin{tabular}{|c|ccc|ccc|}
\hline
 & \multicolumn{3}{c|}{50\%} & \multicolumn{3}{c|}{100\%} \\ \hline
 &
  \multicolumn{1}{c|}{F1 $\uparrow$} &
  \multicolumn{1}{c|}{RPE $\downarrow$} &
  TIME $\downarrow$ &
  \multicolumn{1}{c|}{F1 $\uparrow$} &
  \multicolumn{1}{c|}{RPE $\downarrow$} &
  TIME $\downarrow$ \\ \hline
GNC &
  \multicolumn{1}{c|}{0.747} &
  \multicolumn{1}{c|}{32.20} &
  155.86 &
  \multicolumn{1}{c|}{0.604} &
  \multicolumn{1}{c|}{35.53} &
  349.04 \\ \hline
ADAPT &
  \multicolumn{1}{c|}{0.873} &
  \multicolumn{1}{c|}{1.867} &
  5457.07 &
  \multicolumn{1}{c|}{0.738} &
  \multicolumn{1}{c|}{1.856} &
  12576.27\\ \hline
MAXMIX &
  \multicolumn{1}{c|}{0.77} &
  \multicolumn{1}{c|}{0.324} &
  70.99 &
  \multicolumn{1}{c|}{0.66} &
  \multicolumn{1}{c|}{0.43} &
  137.29\\ \hline
DCS &
  \multicolumn{1}{c|}{0.796} &
  \multicolumn{1}{c|}{26.12} &
  \textbf{4.58} &
  \multicolumn{1}{c|}{0.66} &
  \multicolumn{1}{c|}{37.684} &
  \textbf{8.28}\\ \hline
GM &
  \multicolumn{1}{c|}{0.795} &
  \multicolumn{1}{c|}{23.64} &
  7.07 &
  \multicolumn{1}{c|}{0.65} &
  \multicolumn{1}{c|}{38.33} &
  9.59\\ \hline
HUBER &
  \multicolumn{1}{c|}{0.706} &
  \multicolumn{1}{c|}{21.44} &
  10.04 &
  \multicolumn{1}{c|}{0.55} &
  \multicolumn{1}{c|}{27.32} &
  8.29\\ \hline
PCM &
  \multicolumn{1}{c|}{0.53} &
  \multicolumn{1}{c|}{25.01} &
  8289.04 &
  \multicolumn{1}{c|}{0.371} &
  \multicolumn{1}{c|}{28.54} &
  12655.41\\ \hline
IPC &
  \multicolumn{1}{c|}{\textbf{0.91}} &
  \multicolumn{1}{c|}{\textbf{0.05}} &
  321.37 &
  \multicolumn{1}{c|}{\textbf{0.89}} &
  \multicolumn{1}{c|}{\textbf{0.068}} &
  442.733\\ \hline
\end{tabular}
}
\caption{Summary of the most important metrics in \figref{fig:full_stats} and  in \figref{fig:ate_rpe} for 50\% and 100\% of outliers.}
\label{table:summary}
\vspace{-8mm}
\end{table}

\subsection{Evaluation}
The introduced outliers range from 10\% to 100\% of the ground truth inliers, with increments of ten percent. To ensure fair and robust testing, we generated ten distinct trajectories for each outlier percentage. 
A percentage of 100\% means that the number of outliers is equal to the number of inliers.
As done in \cite{mcgann2023robust}, for methods that don't explicitly classify measurements, an inlier is defined as a measurement having a $\chi^2$ error less than $\chi^2_{th}$ with $\alpha_{th} = 0.95$.
For all methods, the different $CnS$s are evaluated at the end of the optimization of the full problem. 
One small note to take into account is that results for both GNC and PCM were taken after a fixed number of iterations. It was done due to their large convergence time and the exhaustive amount of testing that was carried out for each method.\\
\figref{fig:full_stats} shows the performances for precision, recall, and F1 score averaged on all the used datasets. The precision is one of the most crucial parameters for a correct termination for the optimization. 
Its value should ideally approach one since even minor outlier contributions have the potential to severely compromise the trajectory estimate.
MAXMIX and IPC are the methods that reject most of the outliers across the different datasets, closely followed by PCM. 
Conversely, for GNC, DCS, and GM when the percentage of outliers exceeds 50\%, the precision drops below 0.5, strongly reducing the probability that the solver reaches a good solution.
Simply having high precision is not sufficient, because having poor recall performances will negatively impact the elimination of the accumulated drift due to the high number of inliers discarded, also resulting in poor trajectory estimation.
Still, the acceptable bound is not as strict as the one used for precision. 
As shown in \figref{fig:full_stats}, all methods have an almost constant recall trend as the percentage of outliers increases. GM and DCS demonstrate the highest recall rates, followed by  ADAPT and IPC which provide a consistent recall of over $80\%$.
Nonetheless, IPC outperforms all these methods in the accuracy metric by a large margin, as shown in the RPE column in \tabref{table:summary}. The F1 score, which relates both precision and recall together, is able to give a global view of the actual methods' performances. It can be seen that the methods report a similar trend with respect to the precision. 
It is an expected behavior given the quasi-constant performance of the recall. 
ADAPT slightly dominates for smaller percentages of outliers, while IPC is able to keep stable and reliable results for all the different percentages, resulting in having the best performances for higher percentages of outliers.\\
\figref{fig:ate_rpe} shows respectively the convergence time, the ATE, and the RPE metrics. We can see that for the ATE almost all methods have similar results, with IPC clearly overtaking the other methods. 
ATE highlights how globally brittle the optimization is even in the presence of a small number of outliers, while RPE measures the local consistency of the estimated trajectories.
IPC, MAXMIX, and ADAPT have the best RPE of the different methods.\\
To compare the offline methods' time performances with IPC, it is used the time needed to validate all the loop closure constraints provided in the different datasets. As expected, it is not able to compete with the robust estimators and DCS in terms of speed, but it provides comparable results with more complex methods such as GNC and MAXMIX, even if they were specifically designed to work offline.\\
In \tabref{table:voting_time} for each dataset is reported the Average Convergence Time per Constraint (ACTxC), that is the average convergence time on the subproblem identified by a loop closure constraint. The worst results are the ones for the M3500, because the majority of outliers captured a vast section of the trajectory, leading to predominantly solving subproblems of high dimensionality. Also, in case a false positive of such dimension is accepted, it would negatively impact the clusters' size and the future constraints' convergence time. Still, the results obtained fit the online deployment of the method.

\begin{table}[t]
\centering
\caption{IPC's ACTxC for each dataset as the outlier percentage varies. The reported results are expressed in seconds.}
\resizebox{0.8\linewidth}{!}{
		\begin{tabular}{|c|c|c|c|c|c|}
            \hline
            \multicolumn{1}{|c|}{Dataset}
      & 20\%  & 40\%  & 60\%  & 80\%  & 100\%   \\ \hline
MIT   & 0.053 & 0.050 & 0.050 & 0.050 & 0.044 \\
INTEL & 0.056 & 0.051 & 0.058 & 0.054 & 0.056 \\
M3500 & 0.577 & 0.542 & 0.717 & 0.601 & 0.623 \\
CSAIL & 0.043 & 0.042 & 0.043 & 0.042 & 0.043 \\
FRH   & 0.098 & 0.095 & 0.100 & 0.098 & 0.109 \\
FR079 & 0.042 & 0.042 & 0.043 & 0.041 & 0.043 \\  \hline
        \end{tabular}
        }
	\label{table:voting_time}
 \vspace{-5mm}
\end{table}
\vspace{-1mm}
\section{Conclusions}
\label{sec:conclusions}
\vspace{-1mm}
We presented IPC, a consensus-based method to incrementally build the maximum coherent set of measurements. IPC demonstrates comparable or better performance with respect to recent state-of-the-art approaches while providing online capabilities. In contrast to most methods, IPC shows great generalization capabilities by showing consistent results across the different datasets without changing its parameters, and its only control parameter $s$ is fairly simple to configure. Currently, IPC's primary limitation is its inability to rectify past erroneous decisions. 
Therefore, a possible future direction could involve investigating the specific contexts where IPC is more prone to making errors and also integrating an offline method on top of it that revises and possibly discards the previously integrated measurements, in order to gain the benefits of both methodologies.

\pagebreak

\balance
{
\bibliographystyle{IEEEtran}
\bibliography{references}
}

\end{document}